\documentclass[10pt,twocolumn,letterpaper]{article}
\usepackage{isba}
\usepackage{times}
\usepackage{epsfig}
\usepackage{graphicx}
\usepackage{amsmath}
\usepackage{amssymb}
\usepackage{multirow}
\usepackage{subfigure}
\usepackage[none]{hyphenat}
\usepackage{epsfig}
\usepackage{epstopdf}
\usepackage[hyphens]{url}
\usepackage[breaklinks=true]{hyperref}
\usepackage{breakcites}
\usepackage[numbers,sort&compress]{natbib}
\usepackage{fancyhdr}
\usepackage[margin=0.8in,bottom=0.9in,top=1in]{geometry}
\isbafinalcopy 

\ifisbafinal\fi
\begin{document}

\title{Continuous Authentication Using One-class Classifiers and their Fusion}
\author{Rajesh Kumar\\
Syracuse University\\
New York, USA\\
{\tt\small rkuma102@syr.edu}
\and
Partha Pratim Kundu\\
Nanyang Technological University\\
Singapore\\
{\tt\small partha@ntu.edu.sg}
\and
Vir V. Phoha\\
Syracuse University\\
New York, USA\\
{\tt\small vvphoha@syr.edu}}
\maketitle
\thispagestyle{empty}

\begin{figure}[b]
\parbox{\hsize}{\em
2018 IEEE $4^{th}$ International Conference on Identity, Security, and Behavior Analysis (ISBA) \\
978-1-5386-2248-3/18/\$31.00 \ \copyright 2018 IEEE
}\end{figure}
\begin{abstract}
While developing continuous authentication systems $(CAS)$, we generally assume that samples from both genuine and impostor classes are readily available. However, the assumption may not be true in certain circumstances. Therefore, we explore the possibility of implementing $CAS$ using only genuine samples. Specifically, we investigate the usefulness of four one-class classifiers $OCC$ (elliptic envelope, isolation forest, local outliers factor, and one-class support vector machines) and their fusion. The performance of these classifiers was evaluated on four distinct behavioral biometric datasets, and compared with eight multi-class classifiers ($MCC$). The results demonstrate that if we have sufficient training data from the genuine user the $OCC$, and their fusion can closely match the performance of the majority of $MCC$. Our findings encourage the research community to use $OCC$ in order to build $CAS$ as they do not require knowledge of impostor class during the enrollment process. 
\end{abstract}
\vspace{-0.1in}
\section{Introduction}
\par The multi-class classifiers $(MCC)$ have been widely studied for building behavioral biometric based continuous authentication systems \cite{Touchalytics,SwipingMovementFusion,Abena,PaceIndependentGait}. The disadvantages of using $MCC$ is that they require samples from both genuine and impostor classes to determine respective decision boundaries. In other words, to build an $MCC$-based biometric system, both genuine and impostor samples must be collected \cite{SeveralOneClass}. For training and testing the classification models, researchers have conventionally used samples from other users than the genuine one as impostors \cite{ArmMovement,Touchalytics,GaitCMUBTAS2012}. The assumption that the impostor samples (samples from the other users) are readily available for training $MCC$ might not be realistic in certain circumstances \cite{SmartStrokeOneClassJustificationForOneClass, SeveralOneClass,WittenOCC,PULearning}. For example, (1) individuals might refuse to give their consent for using their biometric data for building authentication systems for someone else, (2) government might pose a restriction on using one's data for building system for others, and (3) the difficulty in collecting good quality impostor samples-- we may have to reveal a great deal of information about the genuine user which could cause privacy concerns \cite{PhoneTyping4}. 
\par Moreover, Manuele et al. \cite{OneClassFace} have demonstrated that the choice of the impostor heavily impacts the performance of $MCC$-based authentication systems. In addition, samples from individuals who were not part of the training database may appear anytime during the verification. This could happen to any realistic (especially adversarial) setup where continuous authentication is applicable. Various studies have advised that even if impostor samples were available, the authentication systems preferably be built by using only genuine samples \cite{WittenOCC,OCCSurvey}. Thus, we explored the possibility of implementing $CAS$ by employing genuine samples only. Specifically, we implemented $CAS$ using four different $OCC$ and their fusion, and tested their performance on four distinct behavioral biometric datasets. The $OCC$ are popular and have been extensively studied for outlier or novelty detection \cite{OCCAnomaly,ZhiruoEnsembleUnsupervised}, as well as in the physiological biometric-based recognition systems \cite{OneClassFace}. However, practicability of $OCC$ has been rarely explored in the context of motion sensor based $CAS$ \cite{MouseDynamics1SVM,WittenOCC}.

\par Ding and Ross applied ensemble of one-class SVMs for detecting spoofing attack on fingerprint recognition system \cite{SppofDetectionWithOCC}. The ensemble of $OCC$ was able to address the insufficient (or unseen) spoof samples problem encountered by conventional spoof detection algorithms. However, the $OCC$ have been rarely explored on motion or touch sensor based biometric datasets. Since these datasets are different in nature and have high variability, it is worth investigating how would the $OCC$ perform on them, especially in the context of continuous authentication. To this end, we hypothesized that if we have sufficient amount of genuine data for training, the $OCC$ could closely match the performance of $MCC$. The main contribution of this paper is summarized below:
\begin{itemize}
\item We investigate four $OCC$ (novelty or outlier detectors) algorithms and their fusion in the context of motion and touch gesture based continuous authentication system. To the best of our knowledge, three of the which have not been studied in continuous authentication domain before.
\item The performance of $OCC$ was compared with eight well established $MCC$ across four distinct behaviometric datasets using False Accept Rate (FAR), False Reject Rate (FRR), Half Total Error Rate (HTER), and Area Under the Curve (AUC). A series of statistical tests were conducted to compare the top performing classifiers from both $MCC$ and $OCC$ groups.
\item The challenges of implementing $CAS$ using $OCC$ including dynamic score normalization in absence of the impostor score distribution is also discussed.
\end{itemize}
\vspace{-0.025in}
\par The rest of the paper is organized as follows: Section \ref{RelatedWork} presents related work; Section \ref{DesignOfExperiments} describes the experimental setup; Section \ref{PerformanceEvaluation} discusses the performance evaluation methods and metrics, and Section \ref{ConclusionAndFutureWork} concludes the work.
\section{Related Work}
\label{RelatedWork}
\par The $OCC$, especially, $SV1C$ have been applied to solve a variety of authentication problems. Examples include face recognition \cite{OneClassFace}, typist recognition \cite{WittenOCC}, smart-stroke \cite{SmartStrokeOneClassJustificationForOneClass}, touch \cite{SeveralOneClass}, and mouse dynamics \cite{MouseDynamics1SVM}. Antal et al. \cite{SeveralOneClass} used four $OCC$ that included Parzen density estimator, the k nearest-neighbor (kNN), Gaussian mixtures method and Support Vector Data Description method in order to build authentication systems based on swipe gestures, however, it was unclear whether their authentication framework was one-time or continuous. Moreover, the swipe gestures and micro-movements of the device were collected in a constrained environment under a very specific scenario -- while responding to psychological questionnaire. Hence, the kNN and Parzen density estimator achieved mean Equal Error Rate (EER) as low as 0.024, and 0.023 after combining the decisions from successive swipe gestures.
\par Antal et al. \cite{SmartStrokeOneClassJustificationForOneClass} also compared $OCC$ and $MCC$ in the context of keystroke-based authentication on mobile devices and demonstrated that $MCC$ outperformed $OCC$ with 4\% of error rate difference. Hempstalk et al. \cite{WittenOCC} combined the density and class probability estimation to improve the classification performance. They also conducted experiments by using the artificially generated impostor samples and pose the question on how the quality and quantity of artificially generated impostor samples may affect the overall performance. Shen et al. \cite{MouseDynamics1SVM} applied SVM-, Neural Network-, and KNN-based $OCC$ on the mouse-usage patterns. They report the Half Total Error Rates (HTER) of $\sim 8\%$, $\sim 15\%$, and $\sim 15\%$ respectively on a dataset of 5550 mouse-operation samples collected from 37 subjects. Also, they strongly argued that \textit{one-class methods are more suitable for user authentication in real-world applications}.
\par However, none of the above papers have studied the classifiers that are studied in this paper with an exception of one-class support vector machines. To the best of our knowledge, the evaluation of $OCC$ has not been done across distinct behavioral biometric datasets before. Moreover, we explore the fusion $OCC$ at score and decision level and discuss the challenges that we faced due to continuous authentication paradigm.  
\section{Design of Experiments}
\label{DesignOfExperiments}
\subsection{Continuous Authentication} 
\par Continuous authentication is a process in which users are unobtrusively monitored at frequent intervals throughout their interaction with any device or system \cite{VishalSurvey}. The $CAS$ pose different challenges compared to the one-time (or login) time authentication systems. Examples include the availability of data throughout the user interaction, high intra-user variance, authentication accuracy, and resource consumption. At the same time, the $CAS$ do not have to be as accurate as the login time authentication systems. Because the verification happens at quite frequent intervals and users can be locked out after certain successive rejects. To implement the continuous part of the system, generally, a sliding window-based mechanism is used \cite{ArmMovement,SwipingMovementFusion,TreadmiLL}. The authentication decisions are given either based on the patterns captured in the current window or in the last few windows. We followed the window-based feature extraction strategy for all four datasets that were studied. The preprocessing, window-size, sliding intervals, and the set of features are kept exactly as advised in the works that have originally proposed the corresponding data set.
\subsection{Datasets and Feature Analysis} 
\par We used four distinct behavioral datasets that included phone-accelerometer based gait patterns, watch-accelerometer based gait, watch-gyroscope based gait, and fusion of swiping and phone movement patterns. These datasets were built with the aim of replicating realistic environment. The training and testing data were collected in separate sessions. The specific details of each dataset are provided below. 
\subsubsection{Phone Acceleration-based Gait Biometric} 
\indent This dataset consists of walking patterns collected through smartphone accelerometer from 18 users who were either faculty, staff or students \cite{TreadmiLL}. Android's \textit{type\_linear\_acceleration} was used that recorded the accelerations (with no gravity component) in the sensor's own frame of reference. The data was collected in two separate sessions, separated by two to three days, referred to as training and testing. The participants walked back and forth freely for about 200 meters keeping HTC One M8 smartphone in their pant pocket. The sampling rate of the accelerometer was set to \textit{normal} which produced around 46 samples per second. The steps of data preprocessing, feature analysis, generation of genuine and impostor samples for training and testing were replicated exactly as advised in \cite{TreadmiLL}. This dataset was the smallest in terms of volume, as the average number of total samples per user per session were 16. This dataset would be referred to as the Phone Acceleration based Gait ($PABG$) in the rest of the paper. 
\subsubsection{Smartwatch-based Gait Biometric} 
\par This dataset contains arm movement patterns of 40 users. The data was collected using the motion sensors (accelerometer, and gyroscope sensors) built into Samsung Galaxy Gear S. The sampling rate for both the sensors were kept to 25Hz. Thirty-four participants were between 20 and 30 years of age, four between 30 and 35, and rest of them were in their 50s. Gender-wise, ten of the subjects were female, while the rest were male. The data was collected for about 2-3 minutes of a walk with watch worn on the wrist. With the 10 seconds of windows with 5 seconds of intervals, the average number of samples per user per session turned out to be 18.4. We replicated the data preprocessing and feature extraction steps as proposed in \cite{ArmMovement}, and used the selected features as advised for creating genuine and impostor samples for training and testing the classifiers. The dataset was divided into two parts based on the type of sensor used for recording the patterns. The arm acceleration pattern recorded through accelerometer sensor will be referred to as Smartwatch Acceleration-based Gait ($WABG$). Similarly, the arm rotation patterns recorded through the smartwatch gyroscope will be referred to as Smartwatch Rotation-based Gait ($WRBG$) in the rest of this paper. 
\begin{figure*}[htp]
\centering
\begin{tabular}{c}
\subfigure[Decision boundaries of $SV1C$, $EE$, $IF$, and $LOF$ on artificially generated Gaussian (uni-modal) data.]{\epsfig{file=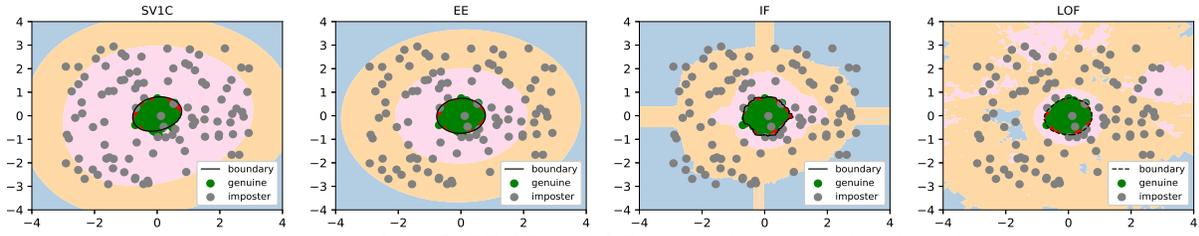, width=6.6in, height=1.25in} 
\label{Unimodal_Boundary}} \\
\subfigure[Decision boundaries of $SV1C$, $EE$, $IF$, and $LOF$ on artificially generated multi-modal data.]{\epsfig{file=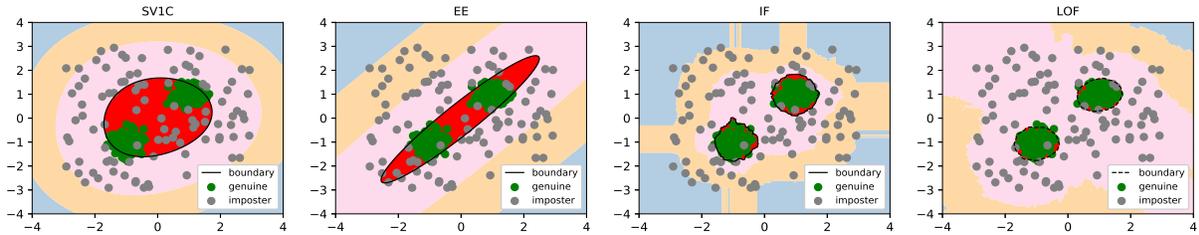, width=6.6in, height=1.25in}
\label{Multimodal_Boundary}}
\end{tabular}
\caption{Illustration of working philosophies of all four $OCC$ that were studied.}
\label{IntuitionForFusion}
\end{figure*}
\subsubsection{Swiping and Phone Movement Patterns} 
\par This dataset consists of swiping patterns along with corresponding underlying phone movement patterns continuously collected through accelerometer sensor while participants browsed specifically designed web pages as well as pages of their choice and answered a set of questions. The data was collected from 28 volunteers in a completely realistic and unconstrained environment for four to seven days. We replicated the data preprocessing and feature extraction steps exactly as proposed Kumar et al. in \cite{SwipingMovementFusion}, and used the selected features as advised for creating genuine and impostor samples. The average number of samples per user was 55.35 and 55.82 for training and testing sessions respectively. This dataset was the biggest in terms of volume. We will refer to this dataset as Swiping and Phone Movement Patterns ($SPMP$) in the rest of the paper. For this dataset, we replicated only one of the experiments i.e., a feature-level fusion of swiping and phone movements with a single-template framework as presented by Kumar et al. \cite{SwipingMovementFusion}. 
\subsection{Choice of Classifiers} 
\par For each dataset, we replicated the experimental setup advised by respective authors in their papers \cite{TreadmiLL},\cite{ArmMovement}, \cite{SwipingMovementFusion}. For the PABG dataset, the authors studied five classifiers: Bayes Network, Logistic Regression, Multilayer Perceptrons, Random Forests, and Support Vector Machines \cite{TreadmiLL}. Similarly, four classifiers (k Nearest Neighbors, Logistic Regression, Multilayer Perceptrons, and Random Forests) were used to implement $CAS$ on $WABG$ and $WRBG$ \cite{ArmMovement}, and two classifiers (kNN with Euclidean, and Random Forests) were used to implement $CAS$ for $SPMP$ \cite{SwipingMovementFusion}. It is generally difficult to know which classifier fits a data set better in advance. So, we decided to study as many (eight) $MCC$ that covered most of the classifiers which were applied on the above datasets in the past. We used Python's sklearn package \cite{sklearn} for running these classifiers. The parameter settings of these algorithms were calibrated to get the best possible performance. 
\par The $MCC$ are well studied in authentication domain and are not the focus of this study, therefore, we do not provide any details on how do they work. However, the $OCC$ that we studied have been rarely explored (except $SV1C$) in this domain, so we briefly discuss their working philosophies in the following paragraphs. The $OCC$ have been successfully applied to solve a variety of one-class problems in the past. The most widely known and established one is one-class Support Vector Machine $(SV1C)$ \cite{SVM1C}. Hence $SV1C$ was an intentional choice in this study. In addition to $SV1C$, we study Elliptic Envelope ($EE$), Isolation Forest ($IF$), and Local Outlier Factor ($LOF$). The reason behind choosing these algorithms was their different working philosophies and distinct decision-making capabilities.
\par The $SV1C$ is an unsupervised method that learns a decision function from the samples supplied to it. The decision function basically is the result of a process that separates all the data points from the origin and maximizes the distance from the hyperplane to the origin. $SV1C$ has two important parameters, $\nu$: determines the upper bound on the fraction of outliers, and allows control to the trade-off between genuine (normal) and impostor (abnormal) predictions, and $\epsilon$: a value used as the stopping tolerance that affects the number of iterations for optimizing the model. We standardized all the features using StandardScaler of Python's sklearn. the StandardScaler standardize features by removing the mean and scaling to unit variance.
\par During enrollment, only genuine samples were supplied to the $SV1C$ while during the verification, both genuine and impostor samples were tested. Based on the learned decision function, $SV1C$ made the decision. In addition to the normal/abnormal decisions, $SV1C$ returned scores associated with the prediction. We used the scores later for score-level fusion. 
\par The Elliptic Envelop ($EE$) \cite{EEnvelop} is a simple outlier detector that assumes the distribution of data as Gaussian. It fits a robust covariance estimate to the supplied data. In other words, it fits an ellipse to the central sample points. The Minimum Covariance Determinant, a robust estimator of covariance was used. The Mahalanobis distances obtained from this estimate was used to derive a measure of abnormality. As $EE$ is very sensitive to the feature dimensions, hence we applied Principle Component Analysis to find out the principal components and supplied only 30\% of the top-ranked components to the $EE$. Unlike, $EE$, $SV1C$ does not assume any parametric form of the data distribution and therefore models the complex shape of the data much better in general.
\par Generally, the outlier detection in high-dimensional space is very challenging. The Isolation forests \cite{RandomForest}, however, does a decent job in such scenario compared to other algorithms e.g. $EE$ that assumes certain underlying distribution (see the rightmost figure of Figure \ref{Multimodal_Boundary}). It basically isolates samples by randomly selecting a feature and then randomly selecting a split point between the maximum and minimum values of the selected feature. This process is repeated recursively and is represented by a tree structure. The number of required partitioning to isolate a sample is the path length from the root node to the terminating node. The averaged path length, over a forest of such random trees, is translated as the measure for making the final decision. The shorter the path the more the abnormality.
\par The Local Outlier Factor ($LOF$) \cite{LOFPaper} is an unsupervised outlier detector which computes the local density deviation of the given sample with respect to its neighbors. The local density is estimated by the typical distance at which a point can be reached from its neighbors. The samples that have a substantially lower density than their neighbors are considered as outliers. The number of neighbors is an important parameter and is generally kept greater than the minimum number of samples that a cluster contains, and smaller than the maximum number of close by samples that could be potential impostors. 
\begin{figure*}[htp]
\centering
\begin{tabular}{cccc}
\subfigure[$PABG$]{\epsfig{file=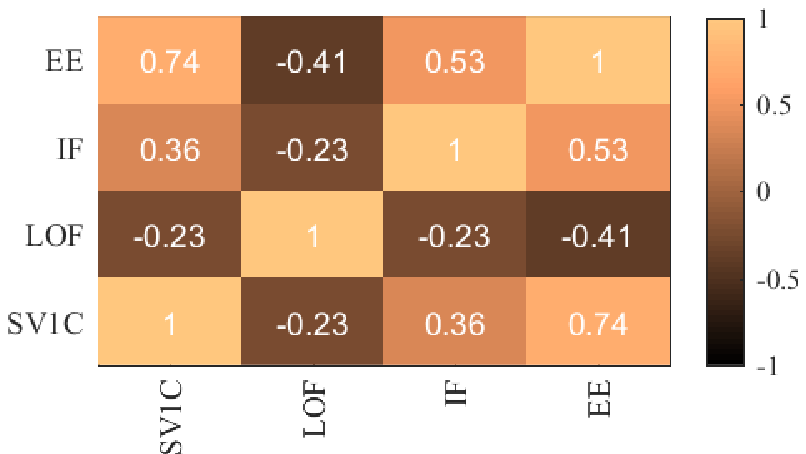,width=1.55in, height=1.2in} 
\label{CorrTreadmill}}
\subfigure[$WABG$]{\epsfig{file=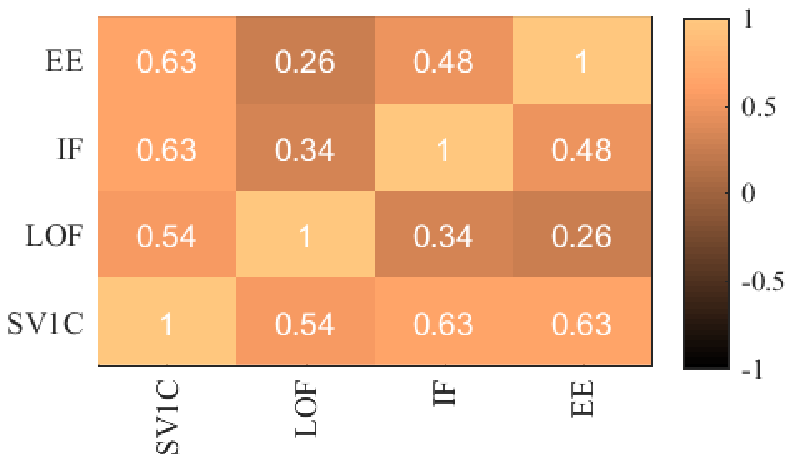,width=1.55in, height=1.2in}
\label{CorrAcc}}
\subfigure[$WRBG$]{\epsfig{file=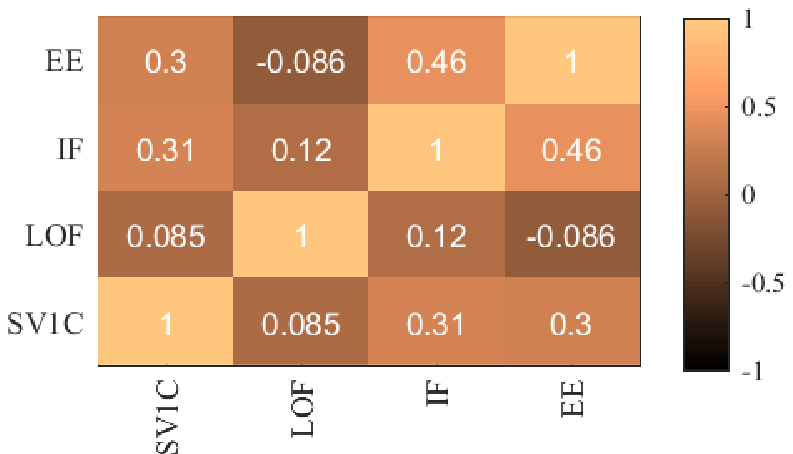,width=1.55in, height=1.2in}
\label{CorrGyro}}
\subfigure[$SPMP$]{\epsfig{file=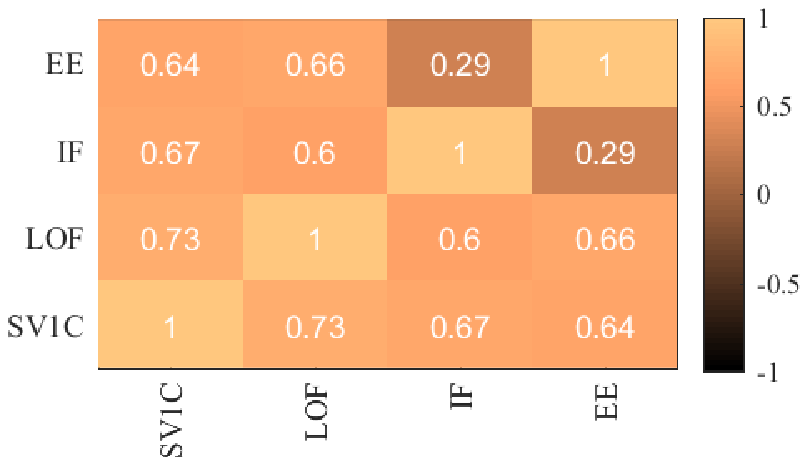,width=1.55in, height=1.2in}
\label{CorrSwipe}}
\end{tabular}
\caption{Heatmap of Pearson correlation coefficients computed among the scores predicted by the OCC in order to gauge the usefulness of the fusion. The less the value of correlation coefficient the more useful fusion might be \cite{ZhiruoEnsembleUnsupervised}.} 
\label{CorrelationAmongClassifiers}
\end{figure*}
\par If the genuine samples from a well-centered elliptical boundary and/or follow a Gaussian distribution, the decision rule based on fitting covariance like $EE$ would be able to generate a well-separated decision boundary around genuine samples. On the contrary, if the genuine samples do not follow any underlying distribution, the $IF$ may perform well (see Figure \ref{Unimodal_Boundary}). Moreover, if the genuine samples are non-Gaussian or multi-modal, $EE$ failed to produce any decision boundary for them but $SV1C$, as well as $IF$, might be able to generate a reasonable decision boundary (see Figure \ref{Multimodal_Boundary}) \cite{OCCComparision}. 
\subsection{Training and Testing of the Classifiers} 
\par Although $OCC$ offer one of the biggest advantages over the $MCC$ i.e. they do not require samples from abnormal (impostor) class at all, they require sufficient training data from the normal (genuine) class for drawing the accurate classification boundary. This phenomenon was observed while setting up the experimental parameters for training verification models as well as user-specific thresholds for authentication systems. 
\par The $MCC$ required both genuine and impostor samples during the training, while \textit{$OCC$ could be trained using genuine samples only}. We tested all ($MCC$ and $OCC$) classifiers for genuine pass/fail rates using genuine samples and impostor pass/fail rates using the impostor samples. For impostor testing, we borrowed a fixed number of samples from other users than the genuine following the suggestions of Kumar et al. \cite{SwipingMovementFusion,TreadmiLL,ArmMovement}. 
\begin{table*}[htp]
\centering
\caption{The performance of $MCC$, $OCC$, and the best fusion of $OCC$ on PABG, WABG, WRBG, and SPMP datasets. The first eight rows (excluding header) presents the average False Accept Rate (FAR), False Reject Rate (FRR), Half Total Error Rates (HTER), and Area Under the Curve (AUC) obtained by $MCC$ on different datasets. The next four rows present the same metrics obtained by four individual $OCC$. Notably, the performance of individual $OCC$, especially, $SV1C$ and $LOF$ is comparable to most of the $MCC$ except the top three i.e. $SVC$, $kNN$ and $LDA$ for SPMP dataset which had good amount of genuine samples (on an average 55 per user) for training the $OCC$. While the performance of other two $OCC$ is poor compared to the top four $MCC$. Although the performance of the top two $OCC$ is not better, they are still better than half of the $MCC$ across all four datasets. Fusion* represents the combination of classifiers that achieved the best error rates. The combination for PABG and WABG was LOF+SV1C, whereas, for WRBG it was IF+LOF+SV1C, and SPMP, it was IF + LOF + SV1C, and EE+LOF+SV1C respectively.}
\vspace{0.05in}    
\label{PerformanceTable}
\scriptsize
\begin{tabular}{|c|c|c|c|c|c|c|c|c|c|c|c|c|c|c|c|c|}
\hline
\multirow{2}{*}{\textbf{Classifier}} & \multicolumn{4}{c|}{\textbf{PABG}}                         & \multicolumn{4}{c|}{\textbf{WABG}}                         & \multicolumn{4}{c|}{\textbf{WRBG}}                         & \multicolumn{4}{c|}{\textbf{SPMP}}                         \\ \cline{2-17} 
                                     & \textbf{FAR} & \textbf{FRR} & \textbf{HTER} & \textbf{AUC} & \textbf{FAR} & \textbf{FRR} & \textbf{HTER} & \textbf{AUC} & \textbf{FAR} & \textbf{FRR} & \textbf{HTER} & \textbf{AUC} & \textbf{FAR} & \textbf{FRR} & \textbf{HTER} & \textbf{AUC} \\ \hline
\textbf{ABoost}                      & 4.58         & 20.27        & 12.42         & 87.58        & 2.24         & 24.34        & 13.29         & 86.71        & 2.56         & 25.38        & 13.97         & 86.03        & 8.33         & 13.41        & 10.87         & 89.13        \\ \hline
\textbf{NBayes}                      & 1.96         & 27.86        & 14.91         & 85.09        & 1.28         & 25.47        & 13.38         & 86.62        & 2.88         & 28.76        & 15.82         & 84.18        & 9.79         & 11.04        & 10.42         & 89.58        \\ \hline
\textbf{kNN}                         & 13.07        & 1.48         & 7.28          & 92.72        & 7.56         & 3.99         & 5.78          & 94.22        & 7.95         & 8.71         & 8.33          & 91.67        & 14.02        & 3.87         & 8.94          & 91.06        \\ \hline
\textbf{LDA}                         & 10.78        & 3.81         & 7.30          & 92.70        & 6.47         & 8.69         & 7.58          & 92.42        & 6.28         & 12.61        & 9.45          & 90.55        & 18.52        & 5.38         & 11.95         & 88.05        \\ \hline
\textbf{LReg}                        & 6.21         & 7.09         & 6.65          & 93.35        & 4.62         & 17.33        & 10.97         & 89.03        & 6.47         & 23.41        & 14.94         & 85.06        & 12.04        & 7.91         & 9.97          & 90.03        \\ \hline
\textbf{MLP}                         & 7.52         & 7.61         & 7.56          & 92.44        & 3.40         & 18.75        & 11.07         & 88.93        & 4.17         & 22.17        & 13.17         & 86.83        & 15.08        & 6.72         & 10.90         & 89.10        \\ \hline
\textbf{RFC}                         & 2.29         & 14.62        & 8.45          & 91.55        & 0.58         & 26.58        & 13.58         & 86.42        & 1.47         & 25.66        & 13.57         & 86.43        & 7.80         & 12.06        & 9.93          & 90.07        \\ \hline
\textbf{SVC}                         & 14.05        & 2.50         & 8.28          & 91.72        & 3.21         & 10.03        & 6.62          & 93.38        & 4.49         & 15.10        & 9.79          & 90.21        & 18.12        & 3.04         & 10.58         & 89.42        \\ \hline
\hline
\textbf{SV1C}                        & 7.03         & 14.65        & 10.84         & 89.16        & 9.01         & 13.01        & 11.01         & 88.99        & 11.83        & 16.45        & 14.14         & 85.86        & 11.71        & 9.48         & 10.59         & 89.41        \\ \hline
\textbf{LOF}                         & 6.70         & 17.78        & 12.24         & 87.76        & 18.46        & 11.72        & 15.09         & 84.91        & 18.75        & 11.86        & 15.31         & 84.69        & 12.83        & 10.89        & 11.86         & 88.14        \\ \hline
\textbf{IF}                          & 17.16        & 25.15        & 21.15         & 78.85        & 16.25        & 22.35        & 19.30         & 80.70        & 12.56        & 21.28        & 16.92         & 83.08        & 15.15        & 24.56        & 19.85         & 80.15        \\ \hline
\textbf{EE}                          & 14.05        & 27.43        & 20.74         & 79.26        & 19.26        & 14.38        & 16.82         & 83.18        & 23.62        & 20.14        & 21.88         & 78.12        & 15.28        & 15.54        & 15.41         & 84.59        \\ \hline
\textbf{Fusion*}                     & 8.17         & 13.61        & 10.89         & 89.11        & 7.37         & 17.29        & 12.33         & 87.67        & 10.58        & 17.83        & 14.20         & 85.80        & 11.51        & 9.24         & 10.37         & 89.63        \\ \hline
\end{tabular}
\end{table*}
\subsection{Fusion of $OCC$} 
\par We explored the fusion of $OCC$ considering the fact that they work on different philosophies and create different decision boundaries (see Figure \ref{IntuitionForFusion}) as well as the relatively low correlation among the scores obtained by the classifiers (see Figure \ref{CorrelationAmongClassifiers}). The fusion of two classifiers could enhance the performance when their decision or scores/decisions are uncorrelated from each other \cite{ScoreNormalization, ZhiruoEnsembleUnsupervised,ScoreNormalization}. There exists several other methods to measure the diversity of classifier for usefulness of the fusion, and their relationship to the overall performance that are discussed by Kuncheva et al. \cite{MeasureOfDiversity}. We aim to explore more $OCC$ and their usefulness in the fusion to enhance the overall performance of the system in future.

\par In our experimental setup, the fusion of the $OCC$ was feasible at both score or decision-levels. The decision-level fusion could not improve the performance of the overall system. Further, we explored the option of training a classifier to fuse the decisions from all four $OCC$. However, the number of decision samples were too low to train a classifier for $PABG$, $WABG$, $WRBG$. So we trained the fusion-classifier for $SPMP$ but observed no improvement in the performance compared to the individual $OCC$. Similarly, we also explored the option of training an $SV1C$ using the scores obtained from all four $OCC$ to carry out score-level fusion, but we observed only minor improvements in the overall performance.

\par One of the biggest challenges that we faced while fusing the score was the normalization of scores obtained by different classifiers on the same scale. In case of $MCC$, the min-max normalization has been an established solution, especially when the distribution of the score distribution is unknown. Essentially we need to know the both genuine and impostor scores to compute the min and max, however, in case of $OCC$ we only know the genuine scores. To generate the impostor scores, one could collect some impostor data, however, that would be against what we are establishing through this paper, i.e. \textit{implementing a continuous authentication system using only genuine samples}. In our experiment, we used the following logistic function $\sigma_n = 1/(1+\exp(-\beta \times \sigma_s))$ with different values of $\beta$ that was decided based on the genuine scores obtained by running a validation on the training set itself. Where $\sigma_s$ is the original score and $\sigma_n$ is the normalized scores. We also tested tanh $\sigma_n = (2/(1+\exp(-2\times\beta\times \sigma_s))-1)$ and soft-sign $(\sigma_n = \sigma_s/1+abs(\sigma_s))$ function they worked fine too. The combined score was evaluated against a threshold (derived from the genuine scores obtained on the training data) to make the final decision. 

\section{Performance Evaluation}
\label{PerformanceEvaluation}
\par
\begin{figure*}[htp]
\centering
\begin{tabular}{cc}
\subfigure[$PABG$]{\epsfig{file=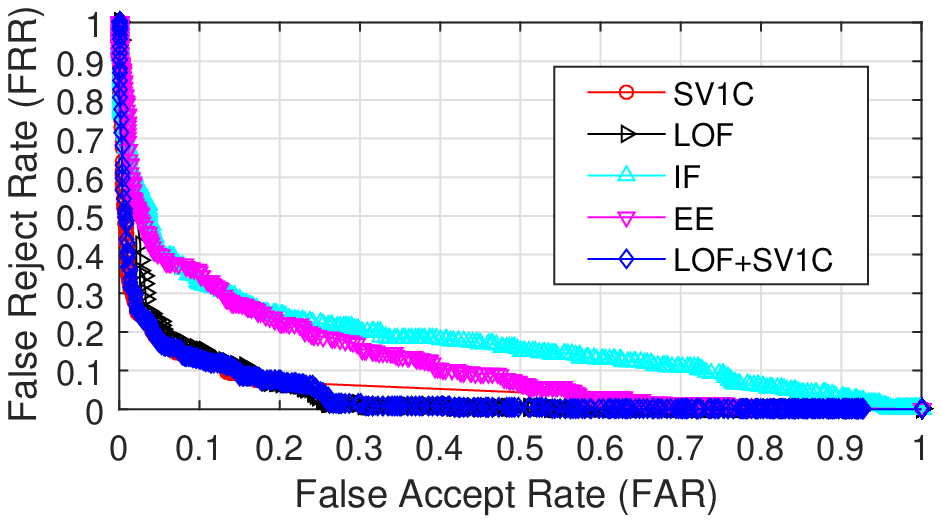,width=3.2in, height=1.6in} 
\label{DETTreadmill}}&
\subfigure[$WABG$]{\epsfig{file=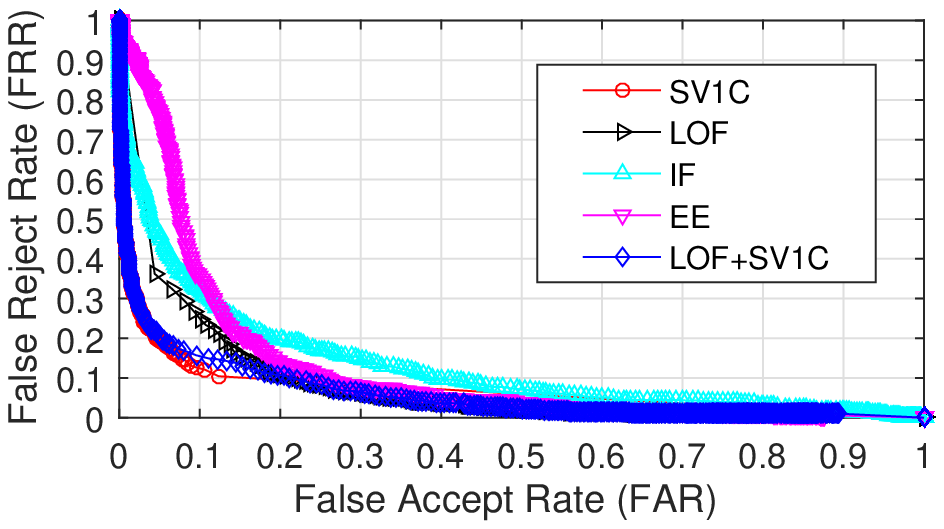,width=3.2in, height=1.6in}
\label{DETGyro}}\\
\subfigure[$WRBG$]{\epsfig{file=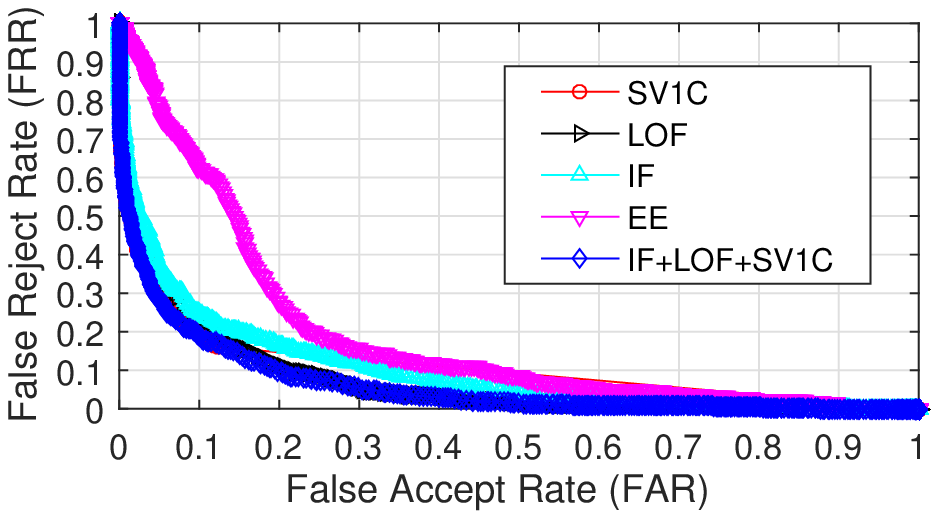,width=3.2in, height=1.6in}
\label{DETAcc}}&
\subfigure[$SPMP$]{\epsfig{file=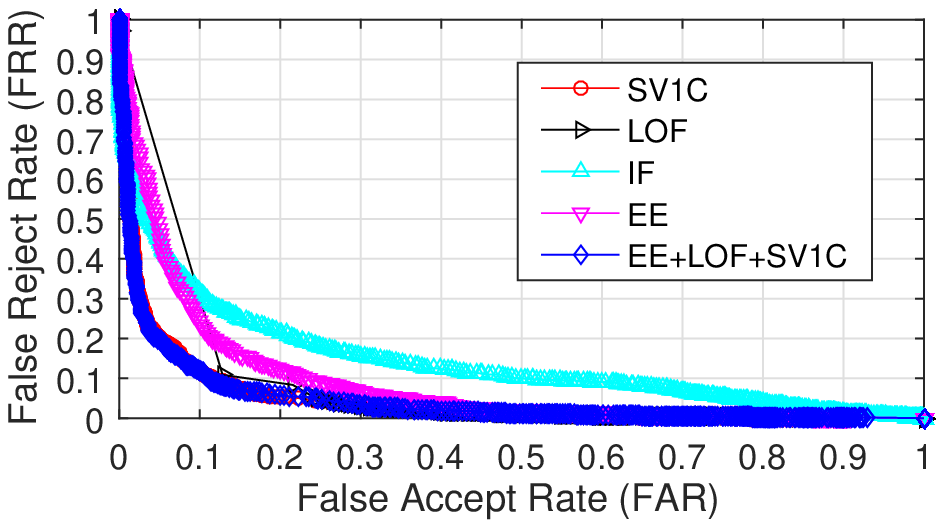,width=3.2in, height=1.6in}
\label{DETSPMP}}
\end{tabular}
\caption{Illustration of trade-off between two error measures False Accept Rates (FAR) and False Reject Rates (FRR). These curves were plotted by varying the decision threshold on the scores obtained by the $OCC$ and the best fusion, computing the mean FAR and FRR at every threshold across the user population of respective datasets.The curve of SPMP looks quite consistent and smooth because we had good amount of training data (56 samples per user) compared to the other datasets where we had around 18 samples per user. Another crucial observation is that $SV1C$ dominated the fusion heavily.}
\label{DETCurveFigure}
\end{figure*}

 The performance of all classifiers was evaluated on four different datasets using false accept rate (FAR), false reject rate (FRR), Half Total Error Rates (HTER), and Area Under the Curve (AUC) \cite{HTERMetric,MouseDynamics1SVM,BalaganiTIFS}. HTER is defined as the average of FAR and FRR. The important hyperparameters of most of the classifiers were calibrated to achieve the best possible error rates. 
To understand the operational characteristic of the $OCC$ and fusion-based systems, we plotted the Detection Error Trade-off (DET) curve \cite{DETCurve, DETCurveJain} (see Figure \ref{DETCurveFigure}). These curves were plotted by varying the decision threshold on the scores and computing the mean FAR and FRR for each threshold across the user population of respective datasets. We observed that $SV1C$, $LOF$, and the fusion is doing significantly well. Also, we can observe that the curve is smooth for the SPMP that had good amount of training samples among all datasets. 
\begin{table*}[htp]
\centering
\caption{The p-Values obtained from the statistical test conducted to evaluate the significance of difference among the error rates (HTER) obtained by the top four $MCC$ and top two $OCC$ for each dataset. Note that the Friedman test, Wilcoxon signed ranked test, kolmogorov smirnov test, and Classification Pair are abbreviated as Fried., Wilc., KST, and CL Pair in this table.}
\vspace{0.02in}
\label{StatisticalTests}
\tiny
\begin{tabular}{|c|c|c|c|c|c|c|c|c|c|c|c|c|c|c|c|}
\hline
 \multicolumn{4}{|c|}{\textbf{PABG}}                 & \multicolumn{4}{c|}{\textbf{WABG}}                & \multicolumn{4}{c|}{\textbf{WRBG}}                & \multicolumn{4}{c|}{\textbf{SPMP}}                \\ \hline 
                                         \textbf{CL Pair} & \textbf{KS} & \textbf{Wilc.} & \textbf{Fried.} & \textbf{CL Pair} &\textbf{KST} & \textbf{Wilc.} & \textbf{Fried.} & \textbf{CL Pair} & \textbf{KST} & \textbf{Wilc.} & \textbf{Fried.} & \textbf{CL Pair} & \textbf{KST} & \textbf{Wilc.} & \textbf{Fried.} \\ \hline

KNN-1CLOF &6.8e-04 &9.7e-01 &4.4e-01&   KNN-1CLOF &4.4e-08 &4.5e-03 &3.3e-01&   KNN-1CLOF &6.7e-08 &1.8e-02 &8.7e-01&   NBayes-1CLOF &5.4e-05 &5.1e-01 &5.5e-01 \\ \hline
LDA-1CLOF &1.5e-03 &6.7e-01 &2.0e-01&   LDA-1CLOF &9.3e-07 &1.2e-01 &6.2e-01&   LDA-1CLOF &6.3e-07 &1.3e-01 &2.5e-01&   KNN-1CLOF &1.8e-05 &4.8e-01 &2.6e-01 \\ \hline
LR-1CLOF &2.3e-03 &6.4e-01 &7.6e-01&    LR-1CLOF &8.9e-06 &7.2e-01 &8.7e-01&    MLP-1CLOF &1.1e-06 &7.7e-01 &1.3e-01&   LR-1CLOF &8.3e-05 &4.1e-01 &4.5e-01 \\ \hline
MLP-1CLOF &1.0e-03 &8.9e-01 &1.1e-01&   SVC-1CLOF &9.5e-07 &6.0e-03 &4.6e-02&   SVC-1CLOF &3.4e-06 &6.5e-02 &4.1e-01&   RFC-1CLOF &4.8e-05 &7.4e-01 &8.5e-01 \\ \hline
KNN-1CSVM &6.8e-04 &5.1e-01 &7.1e-02&   KNN-1CSVM &9.8e-08 &8.5e-02 &1.0e+00&   KNN-1CSVM &6.7e-08 &5.8e-02 &8.7e-01&   NBayes-1CSVM &3.5e-06 &5.9e-03 &1.7e-03 \\ \hline
LDA-1CSVM &2.4e-03 &5.5e-01 &2.0e-01&   LDA-1CSVM &2.5e-06 &4.0e-01 &6.2e-01&   LDA-1CSVM &6.3e-07 &2.1e-01 &6.2e-01&   KNN-1CSVM &1.0e-05 &3.5e-03 &2.5e-03 \\ \hline
LR-1CSVM &2.3e-03 &7.6e-01 &1.3e-01&    LR-1CSVM &8.9e-06 &9.0e-01 &4.0e-01&    MLP-1CSVM &1.1e-06 &8.7e-01 &1.3e-01&   LR-1CSVM &5.5e-06 &1.3e-03 &2.6e-04 \\ \hline
MLP-1CSVM &1.0e-03 &8.0e-01 &1.1e-01&   SVC-1CSVM &9.5e-07 &5.9e-02 &3.0e-01&   SVC-1CSVM &3.4e-06 &9.0e-02 &8.7e-01&   RFC-1CSVM &4.9e-06 &2.2e-02 &1.2e-01 \\ \hline
\end{tabular}
\end{table*}

\par For comparing the performance of the $OCC$ with the $MCC$, we evaluated a total of eight multi-class classifiers: AdaBoost ($ABoost$), Naive Bayes ($NBayes$), k-Nearest Neighbors ($kNN$), Linear Discriminant Analysis ($LDA$), Logistic Regression ($LReg$), Multilayer Perceptron ($MLP$), Random Forest ($RFC$), and Support Vector Classification ($SVC$) over all datasets. The performance of these algorithms is reported in Table \ref{PerformanceTable}. The $kNN$, $LDA$, and $SVC$ achieved the best error rates across the datasets. We can see that $SV1C$ and $LOF$ outperformed at least four $MCC$ across all datasets. All possible (twelve) combinations of four $OCC$ were evaluated. The $SV1C$ dominated across the combinations and dataset. The HTER of the best fusion was either equal to or lower than $SV1C$.
\par A series of statistical tests were conducted in order to ensure if the difference of error rates among the classifiers (1) was not mere fluctuations due to a few extremely good/bad users i.e. -- statistically significant, and (2) holds true for the larger population of users. The tests were conducted pairwise between the top four $MCC$ and top two $OCC$ for each dataset. The top-performing $MCC$ varied across different datasets, however, top performing $OCC$ were consistent across all four datasets. The user-level HTERs were the input to the test methods. The mixed effects Analysis of Variance (MANOVA) method is generally used to test the statistical significance of differences between two samples, in this case, the difference between user-level mean HTERs obtained by the classifiers that were being tested. The MANOVA test assumes that the pairwise difference of mean HTERs across the users follows a Gaussian distribution. To find out if that assumption is true, we applied Kolmogorov-Smirnov (KS) test \cite{KSTest}. The null hypothesis for the KS test was that the difference follows a Gaussian distribution. KS test rejected the null hypothesis at the 5\% significance level for all pair of classifiers that we compared. Table \ref{StatisticalTests} presents the p-values corresponding to KS tests were really low. Hence, the MANOVA was ruled out. 
\par We then used Wilcoxon signed rank test that makes no assumption about the underlying distribution of the error difference. The null hypothesis of this test was that the difference between HTERs achieved by two algorithms was significant and will hold for a larger population. The tests failed to reject the null hypothesis at the 5\% level. We concluded that the difference of error rates holds true for the current as well as a larger population of users. This conclusion was further verified by using Friedman test  \cite{Friedman}. The p-values of the Wilcoxon and Friedman tests are also reported in Table \ref{StatisticalTests}.
\section{Conclusion and Future Work}
\label{ConclusionAndFutureWork}
\par Our findings suggest that it is possible to build behavioral biometrics-based continuous authentication systems without using samples from impostor class. Such systems can be implemented by using $OCC$ and their fusion. The $SV1C$ and $LOF$ achieved comparable error rates and outperformed half of the eight $MCC$. The fusion of $OCC$ could not improve the performance of the system significantly, however, if deeply investigated the fusion would reduce the error rates further. Hence, in the future, we aim to explore the fusion further by considering more $OCC$, and test them on the publicly available behavioral biometric datasets.
\section{Acknowledgement} We thank the anonymous reviewers for their insightful feedback. This work was supported in part by National Science Foundation Award SaTC \#1527795. During the submission of the paper for review, Partha Pratim Kundu was with Syracuse University.
{\small
\bibliographystyle{ieee}
\bibliography{submission_example}

\begin{thebibliography}{33}
\providecommand{\natexlab}[1]{#1}
\providecommand{\url}[1]{\texttt{#1}}
\expandafter\ifx\csname urlstyle\endcsname\relax
  \providecommand{\doi}[1]{doi: #1}\else
  \providecommand{\doi}{doi: \begingroup \urlstyle{rm}\Url}\fi

\bibitem[Frank et~al.(2013)Frank, Biedert, Ma, Martinovic, and
  Song]{Touchalytics}
M.~Frank, R.~Biedert, E.~Ma, I.~Martinovic, and D.~Song.
\newblock Touchalytics: On the applicability of touchscreen input as a
  behavioral biometric for continuous authentication.
\newblock \emph{Trans. Info. For. Sec.}, 8\penalty0 (1):\penalty0 136--148,
  January 2013.
\newblock ISSN 1556-6013.

\bibitem[Kumar et~al.({\natexlab{a}})Kumar, Phoha, and
  Serwadda]{SwipingMovementFusion}
R.~Kumar, V.~V. Phoha, and A.~Serwadda.
\newblock Continuous authentication of smartphone users by fusing typing,
  swiping, and phone movement patterns.
\newblock In \emph{2016 IEEE (BTAS-2016)}, {\natexlab{a}}.

\bibitem[Primo et~al.()Primo, Phoha, Kumar, and Serwadda]{Abena}
A.~Primo, V.~V. Phoha, R.~Kumar, and A.~Serwadda.
\newblock Context-aware active authentication using smartphone accelerometer
  measurements.
\newblock In \emph{CVPRW, 2014}, pages 98--105.

\bibitem[Zhong et~al.(2015)Zhong, Deng, and Meltzner]{PaceIndependentGait}
Y.~Zhong, Y.~Deng, and G.~Meltzner.
\newblock Pace independent mobile gait biometrics.
\newblock In \emph{2015 IEEE 7th International Conference on Biometrics Theory,
  Applications and Systems (BTAS)}, pages 1--8, Sept 2015.
\newblock \doi{10.1109/BTAS.2015.7358784}.

\bibitem[Antal and Szabó(2016)]{SeveralOneClass}
Margit Antal and László~Zsolt Szabó.
\newblock Biometric authentication based on touchscreen swipe patterns.
\newblock \emph{Procedia Technology}, 2016.
\newblock 9th International Conference Interdisciplinarity in Engineering,
  INTER-ENG 2015, 8-9 October 2015, Tirgu Mures, Romania.

\bibitem[Kumar et~al.(2016)Kumar, Phoha, and Raina]{ArmMovement}
R.~Kumar, VV. Phoha, and R.~Raina.
\newblock Authenticating users through their arm movement patterns.
\newblock \emph{CoRR}, abs/1603.02211, 2016.
\newblock URL \url{http://arxiv.org/abs/1603.02211}.

\bibitem[Juefei-Xu et~al.(2012)Juefei-Xu, Bhagavatula, Jaech, Prasad, and
  Savvides]{GaitCMUBTAS2012}
F.~Juefei-Xu, C.~Bhagavatula, A.~Jaech, U.~Prasad, and M.~Savvides.
\newblock Gait-id on the move: Pace independent human identification using cell
  phone accelerometer dynamics.
\newblock In \emph{2012 IEEE Fifth International Conference on Biometrics:
  Theory, Applications and Systems (BTAS)}, Sept 2012.

\bibitem[Antal and Szabó(2015)]{SmartStrokeOneClassJustificationForOneClass}
M.~Antal and L.~Z. Szabó.
\newblock An evaluation of one-class and two-class classification algorithms
  for keystroke dynamics authentication on mobile devices.
\newblock In \emph{20th International Conference on Control Systems and
  Computer Science}, 2015.

\bibitem[Hempstalk et~al.(2008)Hempstalk, Frank, and Witten]{WittenOCC}
Kathryn Hempstalk, Eibe Frank, and Ian~H. Witten.
\newblock One-class classification by combining density and class probability
  estimation.
\newblock ECML PKDD '08, 2008.

\bibitem[Nguyen et~al.(2011)Nguyen, Li, and Ng]{PULearning}
Minh~Nhut Nguyen, Xiao-Li Li, and See-Kiong Ng.
\newblock Positive unlabeled learning for time series classification.
\newblock In \emph{Proceedings of the Twenty-Second International Joint
  Conference on Artificial Intelligence - Volume Volume Two}, IJCAI'11. AAAI
  Press, 2011.

\bibitem[Zheng et~al.(2014)Zheng, Bai, Huang, and Wang]{PhoneTyping4}
Nan Zheng, Kun Bai, Hai Huang, and Haining Wang.
\newblock You are how you touch: User verification on smartphones via tapping
  behaviors.
\newblock In \emph{IEEE-ICNP 2014}, pages 221--232, Oct 2014.
\newblock \doi{10.1109/ICNP.2014.43}.

\bibitem[Bicego et~al.()Bicego, Grosso, and Tistarelli]{OneClassFace}
Manuele Bicego, Enrico Grosso, and Massimo Tistarelli.
\newblock Face authentication using one-class support vector machines.
\newblock In \emph{Proceedings of the 2005 International Conference on Advances
  in Biometric Person Authentication}, IWBRS'05, Berlin, Heidelberg.
  Springer-Verlag.

\bibitem[Khan and Madden(2010)]{OCCSurvey}
Shehroz~S. Khan and Michael~G. Madden.
\newblock \emph{A Survey of Recent Trends in One Class Classification}, pages
  188--197.
\newblock Springer Berlin Heidelberg, Berlin, Heidelberg, 2010.

\bibitem[Rajasegarar et~al.()Rajasegarar, Leckie, Bezdek, and
  Palaniswami]{OCCAnomaly}
Sutharshan Rajasegarar, Christopher Leckie, James~C. Bezdek, and Marimuthu
  Palaniswami.
\newblock Centered hyperspherical and hyperellipsoidal one-class support vector
  machines for anomaly detection in sensor networks.
\newblock \emph{IEEE-TIFS}.

\bibitem[Zhao et~al.(2015)Zhao, Mehrotra, and
  Mohan]{ZhiruoEnsembleUnsupervised}
Zhiruo Zhao, Kishan~G. Mehrotra, and Chilukuri~K. Mohan.
\newblock Ensemble algorithms for unsupervised anomaly detection.
\newblock In \emph{Proceedings of the 28th International Conference on Current
  Approaches in Applied Artificial Intelligence - Volume 9101}, New York, NY,
  USA, 2015. Springer-Verlag New York, Inc.

\bibitem[Shen et~al.(2013)Shen, Cai, Guan, Du, and Maxion]{MouseDynamics1SVM}
C.~Shen, Z.~Cai, X.~Guan, Y.~Du, and R.~A. Maxion.
\newblock User authentication through mouse dynamics.
\newblock \emph{IEEE TIFS}, 2013.

\bibitem[Ding and Ross(2016)]{SppofDetectionWithOCC}
Y.~Ding and A.~Ross.
\newblock An ensemble of one-class svms for fingerprint spoof detection across
  different fabrication materials.
\newblock In \emph{2016 IEEE International Workshop on Information Forensics
  and Security (WIFS)}, pages 1--6, Dec 2016.
\newblock \doi{10.1109/WIFS.2016.7823572}.

\bibitem[Patel et~al.(2016)Patel, Chellappa, Chandra, and
  Barbello]{VishalSurvey}
V.~M. Patel, R.~Chellappa, D.~Chandra, and B.~Barbello.
\newblock Continuous user authentication on mobile devices: Recent progress and
  remaining challenges.
\newblock \emph{IEEE Signal Processing Magazine}, 33\penalty0 (4):\penalty0
  49--61, July 2016.

\bibitem[Kumar et~al.({\natexlab{b}})Kumar, Phoha, and Jain]{TreadmiLL}
R.~Kumar, V.~V. Phoha, and A.~Jain.
\newblock Treadmill attack on gait-based authentication systems.
\newblock In \emph{2015 IEEE (BTAS-2015)}, {\natexlab{b}}.

\bibitem[et~al.(2011)]{sklearn}
Pedregosa et~al.
\newblock Scikit-learn: Machine learning in {P}ython.
\newblock \emph{Journal of Machine Learning Research}, 12:\penalty0 2825--2830,
  2011.

\bibitem[Schölkopf et~al.(2000)Schölkopf, Williamson, Smola, Shawe-Taylor,
  and Platt]{SVM1C}
Bernhard Schölkopf, Robert Williamson, Alex Smola, John Shawe-Taylor, and John
  Platt.
\newblock Support vector method for novelty detection, 2000.

\bibitem[Rousseeuw and Driessen(1999)]{EEnvelop}
Peter~J. Rousseeuw and Katrien~Van Driessen.
\newblock A fast algorithm for the minimum covariance determinant estimator.
\newblock \emph{Technometrics}, 41\penalty0 (3):\penalty0 212--223, August
  1999.
\newblock ISSN 0040-1706.

\bibitem[Liu et~al.(2012)Liu, Ting, and Zhou]{RandomForest}
Fei~Tony Liu, Kai~Ming Ting, and Zhi-Hua Zhou.
\newblock Isolation-based anomaly detection.
\newblock \emph{ACM Trans. Knowl. Discov. Data}, 6\penalty0 (1):\penalty0
  3:1--3:39, March 2012.
\newblock ISSN 1556-4681.

\bibitem[Ankerst et~al.(1999)Ankerst, Breunig, Kriegel, and Sander]{LOFPaper}
Mihael Ankerst, Markus~M. Breunig, Hans-Peter Kriegel, and J\"{o}rg Sander.
\newblock Optics: Ordering points to identify the clustering structure.
\newblock In \emph{Proceedings of the 1999 ACM SIGMOD International Conference
  on Management of Data}, SIGMOD '99, 1999.

\bibitem[learn developers()]{OCCComparision}
Scikit learn developers.
\newblock Novelty and outlier detection.
\newblock \url{http://scikit-learn.org/stable/modules/outlier_detection.html}.

\bibitem[Jain et~al.(2005)Jain, Nandakumar, and Ross]{ScoreNormalization}
Anil Jain, Karthik Nandakumar, and Arun Ross.
\newblock Score normalization in multimodal biometric systems.
\newblock \emph{Pattern Recogn.}, 38\penalty0 (12), December 2005.

\bibitem[Kuncheva and Whitaker()]{MeasureOfDiversity}
Ludmila~I. Kuncheva and Christopher~J. Whitaker.
\newblock Measures of diversity in classifier ensembles and their relationship
  with the ensemble accuracy.
\newblock \emph{Mach. Learn.}

\bibitem[Point(2012)]{HTERMetric}
U.S. Military Academy (USMA) –~West Point.
\newblock Biometrics metrics report v3.0.
\newblock \url{http://www.usma.edu/ietd/docs/BiometricsMetricsReport.pdf},
  2012.

\bibitem[sedenka et~al.(2015)sedenka, Govindarajan, Gasti, and
  Balagani]{BalaganiTIFS}
J.~sedenka, S.~Govindarajan, P.~Gasti, and K.S. Balagani.
\newblock Secure outsourced biometric authentication with performance
  evaluation on smartphones.
\newblock \emph{IEEE-TIFS}, Feb 2015.
\newblock ISSN 1556-6013.

\bibitem[Martin et~al.(1997)Martin, Doddington, Kamm, Ordowski, and
  Przybocki]{DETCurve}
Alvin~F. Martin, George~R. Doddington, Terri Kamm, Mark Ordowski, and Mark~A.
  Przybocki.
\newblock The det curve in assessment of detection task performance.
\newblock ISCA, 1997.

\bibitem[DET(2015)]{DETCurveJain}
\emph{Guidelines for Best Practices in Biometrics Research - MSU CSE}, 2015.
  {IEEE International Conference on Biometrics, {ICB} 2015, Phuket,
  Thailand,19-22 May, 2015}.

\bibitem[Massey(1951)]{KSTest}
Frank~J. Massey.
\newblock The {K}olmogorov-{S}mirnov test for goodness of fit.
\newblock \emph{Journal of the American Statistical Association}, 46\penalty0
  (253):\penalty0 68--78, 1951.

\bibitem[Friedman(1937)]{Friedman}
M.~Friedman.
\newblock The use of ranks to avoid the assumption of normality implicit in the
  analysis of variance.
\newblock \emph{Journal of the American Statistical Association}, 32\penalty0
  (200):\penalty0 675--701, 1937.

\end{thebibliography}
}
\end{document}